%% file: main2.tex
\documentclass[sigconf]{acmart} 
\usepackage{booktabs} 
\usepackage{graphicx}
\usepackage{amsmath}
\usepackage{amssymb}
\usepackage{subcaption} 
\usepackage{enumitem}   
\usepackage{appendix}   

\setcopyright{acmcopyright}
\copyrightyear{2025} 
\acmYear{2025}
\acmDOI{XXXXXXX.XXXXXXX} 

\acmConference[KDD '25]{31st ACM SIGKDD Conference on Knowledge Discovery and Data Mining}{August 03--04, 2025}{Toronto, ON, Canada}
\acmBooktitle{KDD '25: Proceedings of the 31st ACM SIGKDD Conference on Knowledge Discovery and Data Mining, August 03--04, 2025, Toronto, ON, Canada}

\begin{document}
\title{A Tri-Agent Framework for Evaluating and Aligning Question Clarification Capabilities of Large Language Models}

\author{Yikai Zhao}
\affiliation{%
  \institution{Amazon}
  \city{Dallas}
  \state{Texas}
  \country{USA}
}
\email{yikai@amazon.com}

\author{Saurabh Pandey}
\affiliation{%
  \institution{Amazon}
  \city{Seattle}
  \state{Washington}
  \country{USA}
}
\email{saurabpd@amazon.com}

\author{Pradeep Kumar Misra}
\affiliation{%
  \institution{Amazon}
  \city{Dallas}
  \state{Texas}
  \country{USA}
}
\email{pkmisra@amazon.com}

\renewcommand{\shortauthors}{Zhao et al.} 

\begin{abstract}
Large Language Models (LLMs) are increasingly deployed in interactive systems where understanding user intent precisely is paramount. A key capability for such systems is effective question clarification, especially when user queries are ambiguous or underspecified. This paper introduces a novel tri-agent framework for the robust evaluation of an LLM's ability to engage in clarifying dialogue. Our framework comprises three distinct LLM-based agents: (1) a Question Clarifying Agent (QCA), the system under evaluation, tasked with identifying ambiguities and posing clarifying questions; (2) a Respondent Agent (RA), designed to simulate human user responses, potentially including irrelevant or challenging replies; and (3) an Evaluator Agent (EA), an LLM-as-a-judge, which assesses the quality of the dialogue based on a comprehensive set of metrics. We detail a methodology for synthetic data generation in the supply chain domain as an example. We propose metrics evaluating ambiguity handling, question quality, dialogue efficiency, language appropriateness, and final intent alignment. We also briefly discuss the validation of the EA against human judgments. This work provides a structured approach to benchmark, validate, and improve the clarification capabilities of conversational LLM applications.
\end{abstract}




\maketitle

\section{Introduction}
The proliferation of Large Language Models (LLMs) has catalyzed significant advancements in conversational AI \cite{Allouch2021ConversationalAgents}. Systems powered by LLMs are now expected to engage in nuanced, multi-turn dialogues and perform complex tasks based on user instructions \cite{Guan2025EvaluatingLLMAgents}. A critical aspect of effective human-LLM interaction is the model's ability to accurately comprehend user intent, which is often conveyed through queries that may be ambiguous, incomplete, or underspecified \cite{Lyu2024APA}. The capacity of an LLM agent to proactively identify such issues and seek clarification is crucial for task success and user satisfaction \cite{Zhang2024ClarQLLM}. This process typically involves a clarification-rephrasing loop where the agent asks questions to resolve ambiguity and then confirms its refined understanding with the user before proceeding.

Evaluating the efficacy of this clarification process is challenging. Traditional LLM evaluation benchmarks often focus on static input-output tasks \cite{Wang2018GLUE, Wang2019SuperGLUE}, which may not adequately capture the dynamic and interactive nature of clarification dialogues. Evaluating agentic LLMs, which interact with environments or users over multiple steps, requires specialized frameworks \cite{Liu2023AgentBench, Qin2023ToolLLM}. There is a growing need for evaluation frameworks that can specifically assess an agent's ability to manage ambiguity, ask pertinent questions, understand responses, and ultimately derive a clear, actionable user intent within a conversational context \cite{Guan2025EvaluatingLLMAgents}.

This paper proposes a comprehensive framework for evaluating question clarifying agents. Our approach utilizes three LLM-based agents:
\begin{itemize}
    \item \textbf{Question Clarifying Agent (QCA)}: The LLM agent whose clarification capabilities are being evaluated. Its role is to analyze an initial user query, identify missing information or ambiguities, and iteratively ask clarifying questions. It must also handle situations where users provide unsupported entities or dimensions. After getting explicit confirmation from the user about the clarified inquiry, this agent will call a dedicated tool to send the clarified inquiry to corresponding backend for analysis.
    \item \textbf{Respondent Agent (RA)}: An LLM designed to simulate human user responses to the QCA's clarifying questions. The RA can be programmed to provide relevant answers, but also to introduce challenges, such as irrelevant information or ambiguous replies, to test the QCA's robustness and error handling.
    \item \textbf{Evaluator Agent (EA)}: An LLM acting as a "judge." The EA analyzes the complete dialogue transcript between the QCA and RA. It assesses various aspects of the QCA's performance based on predefined metrics, including the quality of clarification, dialogue coherence, and the accuracy of the final interpreted intent. 
\end{itemize}

The core contributions of this paper are:
\begin{enumerate}
    \item A novel tri-agent framework for evaluating the question clarification capabilities of LLMs in an interactive setting.
    \item A methodology for synthetic data generation tailored to the evaluation of clarification dialogues, demonstrated within the supply chain domain.
    \item A suite of evaluation metrics for assessing ambiguity handling, question relevance and timing, language properness, completeness in entity elicitation, and alignment with true user intent.
    \item A discussion on validating the LLM-as-a-judge (EA) against human evaluations and the role of inter-coder reliability in ensuring robust assessment.
\end{enumerate}

We believe this framework offers a scalable and robust approach to benchmark and iteratively improve question clarifying agents, paving the way for more reliable and effective conversational AI systems.

\input{sections/related_work}

\input{sections/eval_workflow}

\input{sections/data_simulation}

\input{sections/experiment_metrics}

\input{sections/results}

\section{Conclusion} 
\label{sec:conclusion_future_work}
This paper introduced a tri-agent framework (QCA, RA, EA) for evaluating LLM question clarification. We outlined synthetic data generation for supply chain, proposed detailed metrics, and emphasized the critical step of validating the LLM-as-a-judge (EA) against human evaluation. This approach enables dynamic, scalable, and validated assessment. This research contributes to building more reliable and user-centric conversational AI systems.

The proposed tri-agent framework offers several benefits for evaluating question clarifying LLMs, including more realistic interactive evaluation, scalability through a validated LLM-as-a-judge, targeted feedback from granular metrics, domain-specific testing via synthetic data, and robustness checks using varied RA personas. However, limitations remain, such as the dependency on the quality and potential biases of the RA and EA, the inherent differences between synthetic and real-world interactions, and the ongoing challenge of perfectly aligning the EA with nuanced human judgment across all subjective metrics.

Despite these considerations, the framework provides a structured and validated pathway to systematically evaluate and iterate on the crucial capability of question clarification in LLMs. Future work will focus on:
\begin{itemize}
    \item Expanding the synthetic dataset with more diverse and complex scenarios, including multi-intent queries and longer dialogues.
    \item Conducting rigorous EA-human alignment studies with multiple annotators and comprehensive ICR analysis across all metrics.
    \item Exploring adaptive RA behaviors that learn to pose more challenging or diverse responses over time.
\end{itemize}

We believe this line of research will contribute to building more intelligent, reliable, and user-centric conversational AI systems capable of truly understanding and responding to user needs.

\begin{acks}
\end{acks}



\bibliographystyle{ACM-Reference-Format}
\bibliography{references} 

\end{document}

%% file: sections/related_work.tex
\section{Related Work}

\subsection{LLM and Agent Evaluation Frameworks}
The evaluation of LLMs and conversational agents is a rapidly evolving field. Traditional metrics like BLEU \cite{Papineni2002Bleu} and ROUGE \cite{Lin2004ROUGE}, while useful for text generation tasks, often fall short in assessing the nuanced aspects of dialogue quality and interaction success \cite{Liu2023GEval}.
Recent research has focused on developing more comprehensive evaluation frameworks. Benchmarks like GLUE \cite{Wang2018GLUE} and SuperGLUE \cite{Wang2019SuperGLUE} assess general language understanding, while BIG-Bench \cite{Srivastava2022BeyondImitation} probes a wider array of capabilities. For conversational agents, evaluation often centers on task completion rates, dialogue efficiency (e.g., number of turns), and user satisfaction \cite{Deriu2021SurveyTOD}. Surveys like the one by Guan et al. \cite{Guan2025EvaluatingLLMAgents} provide a taxonomy of what to evaluate (task completion, response quality, user experience, memory, planning) and how to evaluate (annotation-based, automated metrics, hybrid, self-judging) in multi-turn conversational LLM-based agents. The MINT benchmark focuses on evaluating LLMs in multi-turn interactions involving tools and natural language feedback \cite{Wang2024MINT}. Frameworks like AgentBench \cite{Liu2023AgentBench} and ToolBench \cite{Qin2023ToolLLM} specifically target the evaluation of LLMs as autonomous agents capable of using tools and planning.

\subsection{Evaluating Clarification in Dialogue}
Specifically for clarification, research has explored when and how agents should ask clarifying questions \cite{Aliannejadi2019AskingCQ}. Early work often focused on simpler scenarios. More recent efforts, such as the ClarQ-LLM benchmark \cite{Zhang2024ClarQLLM}, assess an agent's ability to ask questions to resolve uncertainty and gather information in task-oriented dialogues, using metrics like Success Rate and Average Query Discrepancy (AQD). The AGENT-CQ framework also focuses on automatic generation and evaluation of clarifying questions for conversational search \cite{Siro2024AgentCQ}. The importance of ambiguity detection is highlighted in several works, with frameworks like APA (Alignment with Perceived Ambiguity) aiming to guide models to self-disambiguate \cite{Lyu2024APA}. The quality of clarifying questions themselves, in terms of relevance, specificity, clarity, and utility, is crucial. InfoQuest \cite{Krasnashchok2024InfoQuest} provides a benchmark for multi-turn information-seeking dialogues requiring clarification.

\subsection{LLM-as-a-Judge and Human Alignment}
The concept of using an LLM to evaluate the output of another LLM (LLM-as-a-Judge) has gained traction as a scalable alternative to human evaluation \cite{Chiang2023LLMHumanEval, Wang2023IsChatGPTGoodNLG}. Frameworks like G-Eval prompt powerful LLMs with detailed rubrics to grade NLG outputs \cite{Liu2023GEval}. MT-Bench \cite{Zheng2023JudgingLLM} specifically evaluates chat assistants using strong LLMs as judges and demonstrates high agreement with human preferences. While promising, LLM-as-a-Judge systems can exhibit biases (e.g., position bias, verbosity bias) and sensitivity to prompting, necessitating careful validation against human judgments \cite{Wang2023LargeModelsCanSelfCorrect, Shankar2024WhoValidates}. Ensuring the reliability of automated or semi-automated evaluation through comparison with human assessment is critical \cite{EvidentlyAI_LLMJudgeGuide}.

Our proposed framework builds upon these lines of research by integrating an interactive simulation loop with an LLM-as-a-Judge, specifically targeting the multi-faceted problem of question clarification. Unlike static benchmarks, our approach allows for dynamic evaluation of how an agent handles a conversation flow, including unexpected or adversarial responses, within a specific domain context facilitated by synthetic data generation, while also emphasizing the need for validating the judge itself.

%% file: sections/eval_workflow.tex
\section{Proposed Evaluation Framework}
\label{sec:framework}

\subsection{Core Components}

\paragraph{Question Clarifying Agent (QCA)}
The QCA is the primary LLM system whose performance in clarifying user queries is under evaluation. Given an initial user query, which may be ambiguous or incomplete, the QCA's responsibilities include:
\begin{itemize}
    \item \textbf{Ambiguity Detection}: Identifying if the initial query is underspecified, vague, or possesses multiple potential interpretations that hinder direct execution.
    \item \textbf{Clarifying Question Generation}: Formulating relevant, clear, and concise questions aimed at eliciting the specific information needed to resolve ambiguities.
    \item \textbf{Response Understanding}: Interpreting the responses provided by the RA and integrating this new information into its understanding of the user's intent.
    \item \textbf{Iterative Dialogue Management}: Engaging in potentially multiple turns of clarification, maintaining conversational coherence, and adapting its strategy based on the RA's responses.
    \item \textbf{Unsupported Input Handling}: Recognizing when the RA provides entities, dimensions, or intents that are not supported by the underlying system or task, and responding appropriately (e.g., by stating the limitations).
    \item \textbf{Intent Confirmation and Finalization}: Once the QCA deems the intent sufficiently clarified, it formulates a final, well-specified version of the user's question. Ideally, it would also seek confirmation for this understanding.
\end{itemize}

\paragraph{Respondent Agent (RA)}
The RA's role is to simulate a human user interacting with the QCA. It receives clarifying questions from the QCA and generates responses. The RA's behavior is guided by two key inputs in prompts:
\begin{enumerate}
    \item A predefined ground truth well-specified query ($Q_{gt}$), which represents the unambiguous intent the simulated user holds for a given test case. The simulation of this is explained in the next section.
    \item A persona that dictates its response style (e.g., cooperative, prone to providing vague responses, or introducing specific challenges like unsupported intents or entity dimensions).
\end{enumerate}
The RA uses $Q_{gt}$ to ensure its answers are consistent with an underlying goal, while its persona introduces variability and challenges for the QCA.

\paragraph{Evaluator Agent (EA)}
The EA functions as an LLM-as-a-judge. After the interaction between the QCA and RA concludes, the EA is provided with:
\begin{itemize}
    \item The complete dialogue transcript, which includes the initial user query ($Q_{orig}$), each clarifying question from the QCA (denoted $CQ_i$ for turn $i$), each corresponding response from the RA (denoted $R_i$), and the final clarified question proposed by the QCA ($Q_{final}$).
    \item The ground truth well-specified query ($Q_{gt}$) for that test case.
\end{itemize}
Based on this information, the EA assesses the QCA's performance using a predefined set of metrics (detailed in Section \ref{sec:metrics}). The EA outputs quantitative scores and/or qualitative feedback for each metric.

\subsection{Interaction Protocol}
The evaluation process for a single test case unfolds as follows:
\begin{enumerate}
    \item \textbf{Initialization}: A test case is selected from the synthetic dataset. This test case includes an initial (potentially ambiguous) user query ($Q_{orig}$), the corresponding ground truth well-specified query ($Q_{gt}$), and any specific behavioral instructions for the RA.
    \item \textbf{Query Presentation}: The $Q_{orig}$ is presented to the QCA as the starting point of the interaction.
    \item \textbf{Clarification Dialogue}:
        \begin{itemize}
            \item The QCA analyzes $Q_{orig}$ and, if it deems clarification necessary, poses its first clarifying question ($CQ_1$) to the RA.
            \item The RA, guided by $Q_{gt}$ and its assigned persona, generates a response ($R_1$) to $CQ_1$.
            \item This interactive process of the QCA asking a clarifying question ($CQ_i$) and the RA providing a response ($R_i$) continues iteratively.
            \item The dialogue proceeds until the QCA determines it has sufficient information to form a complete and unambiguous understanding of the user's intent.
        \end{itemize}
    \item \textbf{Final Clarified Question Formulation}: The QCA outputs its final understanding of the user's intent as a well-clarified question ($Q_{final}$).
    \item \textbf{Evaluation}: The EA receives the complete dialogue transcript (containing $Q_{orig}, \{CQ_i, R_i\}_{\text{for all turns}}, Q_{final}$) and the $Q_{gt}$. The EA then assesses the QCA's performance based on the metrics outlined in Section \ref{sec:metrics}.
\end{enumerate}

%% file: sections/data_simulation.tex




\section{Synthetic Dataset Generation}
\label{sec:dataset}
To evaluate the QCA in a specific context, we generate a synthetic dataset focused on supply chain queries, a domain rich in multi-entity ambiguities. The dataset construction involves defining base questions, associated entities, and a methodology for generating varied initial user queries ($Q_{orig}$) and their corresponding ground truth well-specified queries ($Q_{gt}$).

\subsection{Base Questions and Entity Specification}
We first define a set of base questions that are inherently ambiguous in a supply chain context. For each base question, we specify a set of possible entities and dimensions (e.g., \texttt{product}, \texttt{site}, \texttt{date}) that could make the question precise. Each entity is also marked as mandatory (M) or optional (O) for a well-formed query in that specific context. A key characteristic of our dataset design is that a fully clarified question ($Q_{gt}$) might not require all optional entities to be filled, reflecting real-world scenarios where users might desire aggregated information or have partial specifications.

\subsection{Generating $Q_{orig}$ and $Q_{gt}$}
For each base question, multiple instances of ($Q_{orig}$, $Q_{gt}$) pairs are constructed.
\begin{itemize}
    \item \textbf{Ground Truth Query ($Q_{gt}$)}: A $Q_{gt}$ is first formulated by instantiating the base question with a specific set of values for its mandatory entities and a subset of its optional entities. This $Q_{gt}$ represents the ideal, unambiguous intent the QCA should eventually recover.
    \item \textbf{Initial User Query ($Q_{orig}$)}: The $Q_{orig}$ is then derived from this $Q_{gt}$ (or the base question directly) by strategically omitting some or all of the specified entity values. This creates variations in the level of ambiguity presented to the QCA:
        \begin{itemize}
            \item $Q_{orig}$ might be the base question itself (maximum ambiguity, no entities specified).
            \item $Q_{orig}$ might include values for some entities (partial ambiguity), but still require clarification for others (including mandatory ones).
            \item $Q_{orig}$ could even include all necessary entity values but be phrased in a slightly ambiguous way or in a manner that still warrants confirmation by a diligent QCA.
        \end{itemize}
\end{itemize}
This process ensures a diverse set of initial queries, challenging the QCA to identify precisely what information is missing or needs confirmation, irrespective of how much information was provided initially. The QCA's task is to determine which entities are already sufficiently clarified in $Q_{orig}$ and which ones require further questioning to align with a potential $Q_{gt}$.

Two examples of base questions are presented below:
\begin{enumerate}[label=\textbf{BQ\arabic*:}, leftmargin=*]
    \item \textbf{Base Query}: "What is the forecast?"
        \textbf{Entities}: Product (O), Site (O), Date (M)
        \textbf{Example $Q_{gt}$}: "What is the sales forecast for product SKU123 at site Warehouse-A for next month?"
        \textbf{Potential $Q_{orig}$ variants derived}:
        \begin{itemize}
            \item "What is the forecast?" (No entities specified)
            \item "What is the forecast for product SKU123?" (Product specified; site and date missing)
            \item "What is the forecast for next month at Warehouse-A?" (Date and site specified; product missing)
        \end{itemize}
    \item \textbf{Base Query}: "Show inventory levels."
        \textbf{Entities}: Product (O), Site (M), Date (M, typically current or specific past date)
        \textbf{Example $Q_{gt}$}: "Show current inventory levels for product category `Electronics' at site Plant-B as of May 15, 2025."
        \textbf{Potential $Q_{orig}$ variants derived}:
        \begin{itemize}
            \item "Show inventory levels." (No entities specified, mandatory ones missing)
            \item "Show inventory levels for Plant-B." (Site specified; date and optional product missing)
            \item "Inventory levels for 'Electronics' as of May 15, 2025." (Product and date specified; mandatory site missing)
        \end{itemize}
\end{enumerate}

\subsection{Configuration of RA}
For each ($Q_{orig}$, $Q_{gt}$) pair, the RA is configured with a persona instructions. The RA's responses are guided by the $Q_{gt}$. To test the QCA's robustness in handling various user inputs, the RA might be instructed to:
\begin{itemize}
    \item Provide specific entity values directly from $Q_{gt}$ when asked.
    \item Respond with "all products" or "not specified for site" if the QCA inquires about an optional entity not present in that specific $Q_{gt}$, or if $Q_{gt}$ implies aggregation. This tests the QCA's ability to correctly interpret and confirm such broad specifiers.
    \item Introduce slight deviations or rephrasing to simulate natural language variability.
    \item Occasionally introduce unrelated entities or intents on purpose to test if QCA can handle the unsupported input.
\end{itemize}
This setup ensures that the QCA is evaluated on its ability to navigate dialogues towards a known ground truth, even when faced with varying degrees of initial ambiguity and diverse user response styles.

%% file: sections/experiment_metrics.tex
\section{Experimental Setup}
\label{sec:experiments}
We instantiated the three agents (QCA, RA, EA) using Claude 3.5 Sonnet. The QCA is the model under evaluation. All agents were purposefully instructed to generate a reasoning within the <think> XML tags before generating their responses within the <output> XML tags. We have found significant performance improvements since we imposed this test-time compute process.


\subsection{Metrics}
\label{sec:metrics}
The EA assesses the Question Clarifying Agent's (QCA) performance based on the full dialogue transcript, the QCA's final proposed clarified question ($Q_{final}$), and the ground truth well-specified query ($Q_{gt}$). The following metrics, adapted from existing literature on LLM and dialogue system evaluation, are employed. Most metrics are scored by the EA on a 1-5 scale (5 being the best), unless otherwise specified. For metrics where clear violations can be identified, such as Completeness of Clarification, each violation results in a 1-point deduction on the scale.

\begin{itemize}
    \item \textbf{Ambiguity Handling (AH)}: Focuses on the QCA's ability to recognize and address initial query ambiguities.
        \begin{itemize}
            \item \textit{Detection Accuracy (AH-DA)}: Evaluates whether the QCA correctly identified that the initial query $Q_{orig}$ was ambiguous or underspecified and required clarification. This is fundamental for initiating the clarification sub-dialogue \cite{Lyu2024APA, Wu2024CLAMBER}. (Binary: Yes/No, or scale if partial ambiguity detection is considered).
            \item \textit{Completeness of Clarification (AH-CC)}: Assesses if the QCA attempted to clarify all necessary missing entities and dimensions as implied by $Q_{gt}$ to ensure the query becomes actionable and well-defined.
        \end{itemize}
    \item \textbf{Question Quality (QQ)}: Pertains to the characteristics of the clarifying questions posed by the QCA.
        \begin{itemize}
            \item \textit{Relevance (QQ-Rel)}: Measures if each clarifying question was directly relevant to resolving the identified ambiguities in $Q_{orig}$ or subsequent user responses, targeting the actual source of confusion \cite{Aliannejadi2019AskingCQ}.
            \item \textit{Clarity \& Conciseness (QQ-CC)}: Assesses if the QCA's questions were phrased in a clear, unambiguous, and brief manner, facilitating easy understanding by the user. Conciseness is also a factor in benchmarks like ClarQ-LLM, which measure query length \cite{Zhang2024ClarQLLM}.
        \end{itemize}
    \item \textbf{Dialogue Efficiency (DE)}: Concerns the efficiency of the clarification interaction.
        \begin{itemize}
            \item \textit{Number of Turns (DE-Turns)}: The total count of QCA-RA conversational turns taken to reach $Q_{final}$. Generally, fewer turns for a successful clarification indicate higher efficiency, a common metric in dialogue system evaluations \cite{Guan2025EvaluatingLLMAgents}.
        \end{itemize}
    \item \textbf{Language Appropriateness (LA)}: Focuses on specific functional language use.
        \begin{itemize}
            \item \textit{Handling of Unsupported Inputs (LA-Uns)}: Measures how effectively and appropriately the QCA managed scenarios where the RA provided entities or dimensions that are not supported by the underlying system (e.g., by clearly stating non-support).
        \end{itemize}
    \item \textbf{Final Question Alignment (FQA)}: Assesses the quality of the QCA's final formulated question ($Q_{final}$) relative to the ground truth ($Q_{gt}$).
        \begin{itemize}
            \item \textit{Semantic Fidelity (FQA-SF)}: Determines how closely $Q_{final}$ matches the meaning and intent of $Q_{gt}$, without distortion or hallucination. This can be supplemented by automated scores like BERTScore \cite{Zhang2019BERTScore} or assessed by an LLM-as-a-judge for faithfulness \cite{Liu2023GEval}. This study adopts the LLM-as-a-judge approach.
            \item \textit{Precision (FQA-Prec)}: Ensures that $Q_{final}$ does not introduce extraneous or incorrect entities/values that were not part of $Q_{gt}$ or explicitly negotiated and accepted during the dialogue.
        \end{itemize}
    \item \textbf{Overall Task Success (OTS)}: A holistic binary (Yes/No) or probabilistic measure indicating whether the QCA successfully transformed the ambiguous $Q_{orig}$ into a $Q_{final}$ that is semantically equivalent to, and an actionable specification of, the ground truth $Q_{gt}$. This is analogous to Task Completion Rate in task-oriented dialogue systems \cite{Guan2025EvaluatingLLMAgents} and the Success Rate in benchmarks like ClarQ-LLM \cite{Zhang2024ClarQLLM}.
\end{itemize}
The EA is prompted with these metric dimensions and their respective scales to score each dialogue. For metrics like FQA-SF, automated scores can also supplement the EA's judgment to provide a multi-faceted evaluation.

%% file: sections/results.tex
\section{Results and Analysis}
\label{sec:results}
The evaluation of the QCA was conducted using the tri-agent framework. The synthetic dataset comprised 200 unique dialogues in the supply chain domain. 

The evaluation of the QCA is contingent upon the RA performing as expected, adhering to its persona and the underlying $Q_{gt}$ for each test case. To ensure the RA's reliability, we conducted a small pilot study ($N=20$ dialogues) where the QCA and RA interacted. During this pilot, we monitored the RA's responses for consistency with its instructions and alignment with the intended conversational flow for each scenario. Based on these observations, the RA's system prompts and persona instructions were iteratively refined. The final version of the RA, demonstrating consistent and appropriate behavior according to our requirements, was then used for the full-scale evaluation of QCA reported here. This iterative refinement process for the simulator agent is crucial for the validity of the overall evaluation framework.

The EA assessed QCA's performance across all dialogues. The aggregated results are presented in Table \ref{tab:qca_performance_results}.

\begin{table*}[ht]
  \centering
  \caption{QCA Performance Evaluation Results (Average Scores over 200 dialogues)}
  \label{tab:qca_performance_results}
  \begin{tabular}{@{}llc@{}}
    \toprule
    \textbf{Category} & \textbf{Metric} & \textbf{Score / Value} \\
    \midrule
    Ambiguity Handling (AH) & Detection Accuracy (AH-DA) & 0.92 (Accuracy) \\
                              & Completeness of Clarification (AH-CC) & 4.15 / 5 \\
    \addlinespace
    Question Quality (QQ)     & Relevance (QQ-Rel) & 4.48 / 5 \\
                              & Clarity \& Conciseness (QQ-CC) & 4.25 / 5 \\
    \addlinespace
    Dialogue Efficiency (DE)  & Avg. Number of Turns (DE-Turns) & 4.83 turns \\
    \addlinespace
    Language Appropriateness (LA) & Handling of Unsupported Inputs (LA-Uns) & 3.75 / 5 \\
    \addlinespace
    Final Question Alignment (FQA) & Semantic Fidelity (FQA-SF) & 4.38 / 5 \\
                                 & Precision (FQA-Prec) & 4.29 / 5 \\
    \addlinespace
    Overall Task Success (OTS) & OTS Rate & 0.87 (Success Rate) \\
    \bottomrule
  \end{tabular}
\end{table*}

QCA demonstrated strong capabilities in several areas. The high Ambiguity Detection Accuracy (AH-DA: 0.92) indicated that the agent was proficient at identifying when initial user queries required clarification. Question Relevance (QQ-Rel: 4.48) is also a notable strength. The average number of turns (DE-Turns: 4.83) for clarification was reasonably low, contributing to efficient interactions when successful.


However, the evaluation also highlighted areas for improvement. The score for Handling of Unsupported Inputs (LA-Uns: 3.75) suggested that QCA faced challenges when the RA provided entities or dimensions outside the predefined supported scope. For example, if the RA, in response to a query about ``product type'' for a forecast, mentioned an unsupported category like ``experimental raw materials'' instead of established categories such as ``finished goods'' or ``components'', QCA did not always gracefully guide the conversation back. Instead of clearly stating, ``Forecasting for `experimental raw materials' is not currently supported. Would you like to proceed with 'finished goods' or 'components'?'' it might respond ambiguously or attempt to process the unsupported input, leading to downstream errors. This indicates that additional effort in prompt tuning or providing more in-context learning examples is needed in order to improve the QCA on its handling of such unsupported situations.

The Overall Task Success (OTS) rate of 87\% is promising, yet an analysis of the \textasciitilde 13\% failure cases provides valuable insights into QCA's limitations. These failures often correlated with lower scores in Completeness of Clarification (AH-CC: 4.15). One observed pattern leading to OTS failure occurred when the QCA did not sufficiently reconcile the scope of the final clarified question ($Q_{final}$) with the specific granularity of the ground truth ($Q_{gt}$). For instance, even if the RA responded with a general term like "all products" (which might be a valid intent for that specific dialogue instance if $Q_{gt}$ implied aggregation), QCA sometimes failed to then ensure all other mandatory entities as defined by $Q_{gt}$ were adequately specified. This could lead to a $Q_{final}$ that, while correctly capturing the "all products" aspect, missed a critical mandatory dimension like a specific date range required by $Q_{gt}$, thus being incomplete for backend execution. Another distinct failure pattern involved the QCA prematurely concluding the clarification phase. For example, if $Q_{gt}$ required a specific date range (e.g., "June 2025") and the QCA asked, "For which month are you interested?", the RA might respond, "The upcoming summer month." A robust QCA should recognize "upcoming summer month" as still ambiguous (as it could be June, July, or August) and ask for further specification. However, QCA occasionally accepted such partially resolved responses, leading to a $Q_{final}$ that lacked the necessary precision (e.g., "forecast for the upcoming summer month" instead of "forecast for June 2025"), thereby failing the OTS if the $Q_{gt}$ was more specific. This was particularly noted when the RA's responses were designed to be slightly evasive or incomplete, testing the QCA's persistence in seeking full clarity on all essential parameters required by the underlying task's ground truth specification.

These findings offer critical insights for future development. The QCA's occasional shortcomings in handling unsupported inputs and its tendency towards premature clarification with ambiguous RA responses underscore the need for more sophisticated dialogue management strategies. Specifically, future iterations should focus on enhancing the QCA's ability to: (1) provide constructive alternatives when faced with unsupported requests, rather than simply stating a limitation or attempting to process invalid input; and (2) exhibit greater persistence in probing user responses that are not fully resolved, particularly when such responses could lead to an incomplete or imprecise final query relative to downstream task requirements. Improving these aspects is key to not only increasing the OTS rate but also enhancing the overall robustness and user-friendliness of the clarification agent. This points towards the necessity of prompt tuning for the QCA with a diverse range of challenging conversational scenarios, including those that explicitly test boundary condition handling and deep ambiguity resolution.

\subsection{EA-Human Alignment: Pilot Study and Future Directions}
\label{subsec:ea_human_alignment_pilot}
Validating the EA's judgments against human assessment is crucial for establishing the reliability of the LLM-as-a-judge paradigm \cite{Zheng2023JudgingLLM}. To this end, we conducted a pilot study focusing on a key metric: Completeness of Clarification (AH-CC). A randomly selected subset of 50 dialogues was annotated by the author, who was thoroughly familiar with the evaluation rubric. The EA's scores for AH-CC on these 50 dialogues were then compared to the author's scores. This pilot revealed a Pearson's correlation coefficient (r) of $0.87$ between the EA's and the author's scores for the AH-CC metric. This indicates a good level of agreement and suggests that the EA is capable of assessing the completeness of clarification with a reasonable degree of alignment to human judgment for this particular metric. 

However, this single-annotator, single-metric pilot study is a preliminary step. For a more comprehensive validation of the EA across all metrics and to ensure the robustness of the human baseline, a scaled human evaluation study is essential for future work. This would involve:
\begin{enumerate}
    \item \textbf{Multiple Annotators}: Employing multiple independent human annotators trained on the complete evaluation rubric.
    \item \textbf{Annotator Training and Calibration}: Conducting thorough training sessions and calibration exercises to ensure all annotators share a consistent understanding of each metric and scoring scale.
    \item \textbf{Inter-Coder Reliability (ICR) Calculation}: Measuring the consistency among the human annotators using established metrics such as Krippendorff's Alpha ($\alpha$) \cite{Krippendorff2004ContentAnalysis} or Cohen's Kappa (for categorical ratings) \cite{Artstein2008InterCoder}. Achieving satisfactory ICR (e.g., $\alpha > 0.70$) is critical to establish a reliable human gold standard.

\end{enumerate}
This scaled approach is necessary to definitively establish the EA's reliability and identify specific areas where its judgment may diverge from nuanced human perception, thereby guiding further refinement of the EA's prompting.